\documentclass[10pt,twocolumn,letterpaper]{article}

\usepackage{iccv}
\usepackage{times}
\usepackage{epsfig}
\usepackage{graphicx}
\usepackage{amsmath}
\usepackage{amssymb}
\usepackage{multirow}

\usepackage[pagebackref=true,breaklinks=true,letterpaper=true,colorlinks,bookmarks=false]{hyperref}

\iccvfinalcopy 

\def\iccvPaperID{728} 

\ificcvfinal\fi
\begin{document}

\title{Robust High Quality Image Guided Depth Upsampling}

\author{Wei Liu$^{1,2}$, Yijun Li$^{1,2}$, Xiaogang Chen$^3$, Jie Yang$^{1,2}$, Qiang Wu$^{4}$ and Jingyi Yu$^{5}$\\
$^1$Shanghai Jiao Tong University, Shanghai, China. {\tt\small \{liuwei.1989,leexiaoju,jieyang\}@sjtu.edu.cn}\\
$^2$Key Laboratory of System Control and Information Processing, Ministry of Education, China.\\
$^3$University of Shanghai for Science and Technology, Shanghai, China. {\tt\small xg.chen@live.com}\\
$^4$University of Technology, Sydney, Australia. {\tt\small Qiang.Wu@uts.edu.au}\\
$^5$University of Delaware, Newark, DE, USA. {\tt\small yu@cis.udel.edu}
}

\maketitle

\begin{abstract}

Time-of-Flight (ToF) depth sensing camera is able to obtain depth maps at a high frame rate. However, its low resolution and sensitivity to the noise are always a concern. A popular solution is upsampling the obtained noisy low resolution depth map with the guidance of the companion high resolution color image. However, due to the constrains in the existing upsampling models, the high resolution depth map obtained in such way may suffer from either texture copy artifacts or blur of depth discontinuity. In this paper, a novel optimization framework is proposed with the brand new data term and smoothness term. The comprehensive experiments using both synthetic data and real data show that the proposed method well tackles the problem of texture copy artifacts and blur of depth discontinuity. It also demonstrates sufficient robustness to the noise. Moreover, a data driven scheme is proposed to adaptively estimate the parameter in the upsampling optimization framework. The encouraging performance is maintained even in the case of large upsampling e.g. $8\times$ and $16\times$.

\end{abstract}

\section{Introduction}
\label{SecIntroduction}

\begin{figure}
\centering
  \includegraphics[width=1\linewidth]{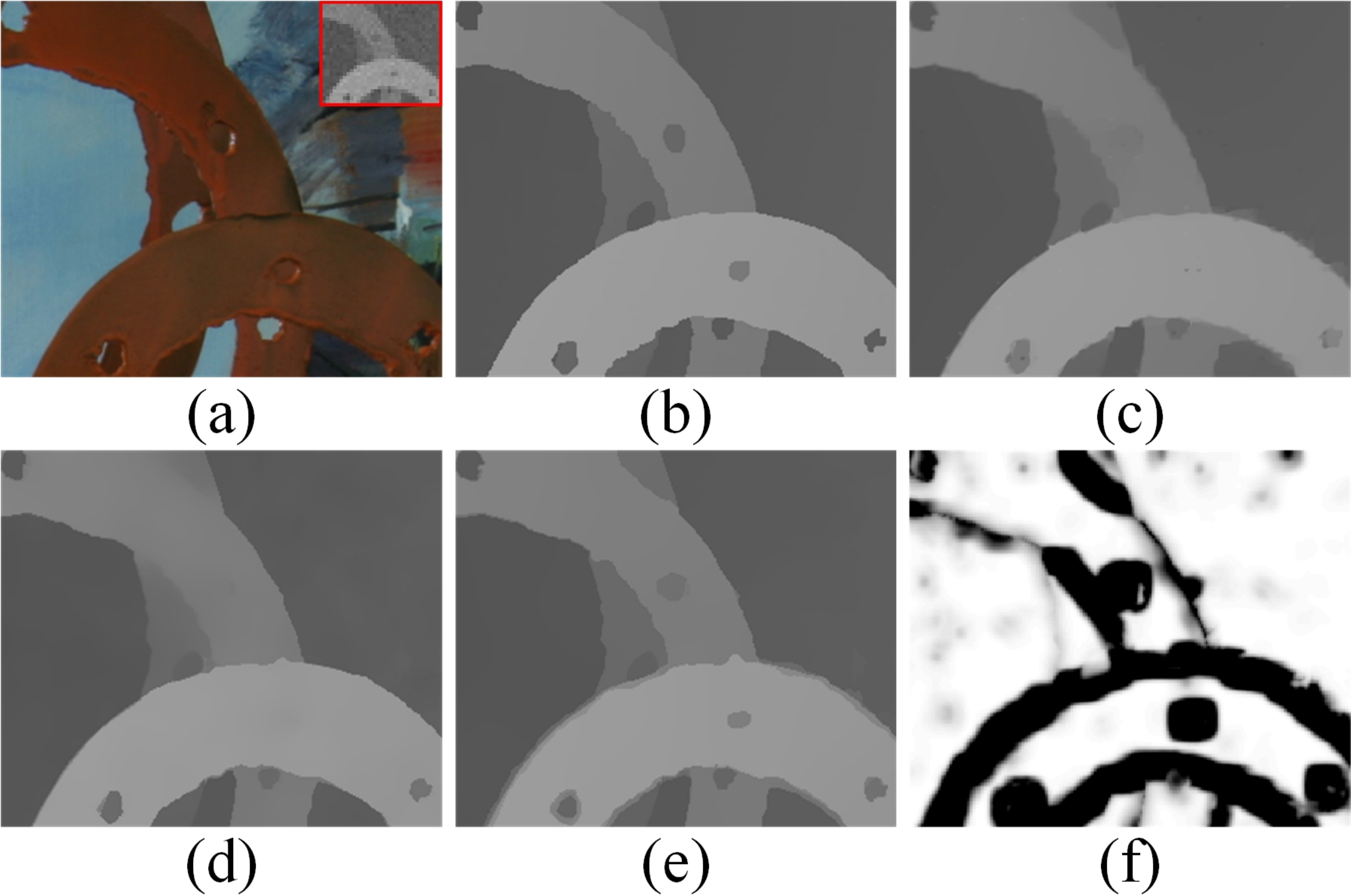}\\
  \caption{$8\times$ upsampling result patches of \emph{Art} from the Middleburry dataset \cite{scharstein2007learning}. (a) The noisy low resolution depth map patch and the corresponding color image. (b) The ground truth. (c) The upsampling result by the state-of-art method in \cite{ferstl2013image}. The upsampling result by our method (d) without adaptive bandwidth selection and (e) with adaptive bandwidth selection. (f) The corresponding bandwidth map obtained by our method.}\label{CoverFigure}
\end{figure}

Acquiring the depth information of 3D scenes is essential for many applications in computer vision and graphics such as 3D modeling, 3DTV and augmented reality. Recently Time-of-Flight (ToF) cameras have shown impressive results and become more and more popular, e.g., Kinect v2.0 sensor. They can obtain dense depth measurements at a high frame rate. However, the resolution is generally very low and the depth map is often corrupted by strong noise.

Tremendous efforts have been put for improving the resolution of depth maps acquired by ToF cameras. The solutions usually go to three categories. In the first category, single low resolution depth map is upsampled through different data-driven approaches. It may be achieved by exploiting similarity across relevant depth patches in the same depth map  \cite{hornacek2013depth}. The resolution can also be synthetically increased by exploiting an existing generic database containing large numbers of high resolution depth patches \cite{li2014similarity}\cite{mac2012patch} which were inspired by the work in \cite{freeman2011markov}\cite{freeman2000learning}.  The second approach upsamples the low resolution depth map by integrating multiple low resolution depth maps of the same scene, which may be acquired at the same time but from slightly different views \cite{rajagopalan2008resolution}\cite{schuon2009lidarboost} or at the different sensing time \cite{hahne2011exposure}. These existing methods assume that the scene is static. The third category achieves upsampling through the supports of the high resolution guided color image \cite{diebel2005application}\cite{ferstl2013image}\cite{kopf2007joint}\cite{min2012depth}\cite{park2011high}\cite{yang2014depth_recovery}\cite{yang2007spatial}\cite{zhu2008fusion}. In this category, it is assumed that the depth discontinuity on the depth map and the color edge on the color image co-occur on the corresponding regions.

Image guided upsampling methods have several advantages against the other two categories and are popular in recent years. They can yield much better upsampling quality than single depth map upsampling \cite{hornacek2013depth}. Besides, they can achieve larger upsampling factor and do not need any prior database compared with the existing methods \cite{mac2012patch}\cite{li2014similarity}. Also when compared with the second category, image guided upsampling is not subject to static scene and does not need complicated camera calibration process. Despite the obvious advantages against the first two categories of the solutions, the main issues of image guided upsampling are: 1) texture copy artifacts on the smooth depth region when the corresponding color image is highly textured; 2) blurring depth discontinuity when the corresponding color image is more homogeneous; 3) performance drops for the case of highly noisy depth maps.

In this paper, a new image guided upsampling approach is proposed. It well tackles the issues of the existing methods in the same category mentioned above. We formulate a novel optimization framework with the brand new data term and the new smoothness term compared with other state-of-art models \cite{diebel2005application}\cite{ferstl2013image}\cite{park2011high}\cite{yang2014depth_recovery}. Pixel-by-patch validity checking is introduced in the data term in the optimization process instead of pixel-by-pixel checking in the existing methods. Also, a new error norm is proposed to replace the L2 norm. These together improve the robustness of the framework against the noise. Moreover, the new smoothness term relaxes the strict assumption on the co-occurrence of depth discontinuity and the color edge which is one of major problems in the existing methods of the same category. To further improve the performance of the proposed framework, we propose a new data driven parameter selection scheme to adaptively estimate the parameter in the optimization process. Experimental results on synthetic and real data show that our method outperforms other state-of-art methods in both visual quality and quantitative accuracy measurement. Encouraging performance retains even for larger upsamping scale such as $8\times$ and $16\times$.

The rest of this paper is organized as follows: Section \ref{SecRelatedWork} is the related work where we briefly present the MRF framework in \cite{diebel2005application} and its extension in \cite{park2011high} which are highly related to our work and then analyze their shortages which motivate the proposed method. In Section \ref{SecOurMethod}, we first detail our upsampling model. Then, we give further explanation to show how the newly proposed model is able to handle the constrains mentioned above. In the end, a data driven parameter estimation scheme is proposed to adaptively select the parameter in the model. Section \ref{SecExperiments} shows our experimental results and the comparison with other state-of-art methods. We draw the conclusion in Section \ref{SecConclusion}.

\section{Related work}
\label{SecRelatedWork}

\begin{figure*}
\centering
  \includegraphics[width=1\linewidth]{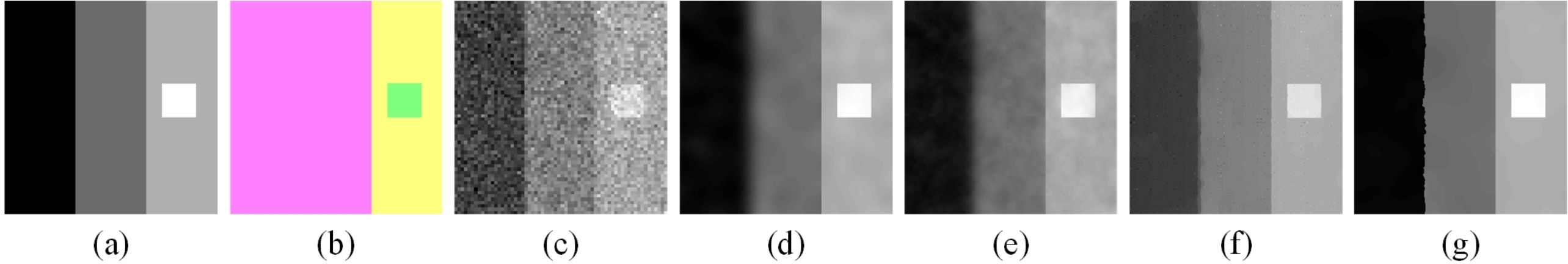}\\
  \caption{(a) The synthetic depth map. (b) Its corresponding synthetic color image and (c) the noisy low resolution depth map (shown in bicubic interpolation). The result obtained by (d) the MRF in \cite{diebel2005application} (RMSE: 1.83), (e) its extension non-local means MRF in \cite{park2011high} (RMSE: 2.07), (f) the image guided total generalized variation upsampling in \cite{ferstl2013image} (RMSE: 1.69) and (g) our method (RMSE: 0.78). (d)$\thicksim$(f) cannot well smooth the noise and properly preserve the depth discontinuity once it does not corresponds to the color edge.}\label{SyntheticComparison}
\end{figure*}

One of the most well known image guided upsampling methods is based on the Markov Random Fields (MRF) \cite{diebel2005application}. It is further extended in \cite{park2011high} which we denote as NLM-MRF. Both methods along with other similar methods such as \cite{ferstl2013image}\cite{kopf2007joint}\cite{liu2013joint}\cite{lu2012cross}\cite{min2012depth}\cite{yang2014depth_recovery}\cite{yang2007spatial} which adopt the certain cues on the corresponding available color image as the reference to upsample the low resolution depth map. They all assume that the color edge co-occurs with the depth discontinuity. To clearly demonstrate the essential constrains of the existing methods, we brief the existing modeling as follows.

Given a noisy low resolution depth map $D^L$ and the companion high resolution color image $I$, the task is to upsample the $D^L$ to $D^H$ of which the resolution is the same as that of $I$. The work in \cite{diebel2005application} is modeled through MRF as\footnote{We have slightly adjusted the original model to have a better comparison with our model. However, the mechanism remains the same.}:
\begin{footnotesize}
\begin{equation}\label{MRF}
    {D}^{H}=\underset{D}{\mathop{\arg\min}}\{(1-\alpha)\sum\limits_{i\in \Omega}{({D}_{i}-D_{i}^{0})^{2}}
    +\alpha \sum\limits_{i\in \Omega }{\sum\limits_{j\in N(i)}{\omega _{i,j}^{c}{({D}_{i}-{D}_{j})^2}}}\}
\end{equation}
\end{footnotesize}
where $D^0$ is the initial value of $D^H$, which can be obtained by interpolating $D^L$ through the basic interpolation such as bicubic interpolation \cite{diebel2005application}. $\Omega$ represents the coordinate of the high resolution depth map. $N(i)$ is the neighborhoods of $i$ in the square patch centered at $i$. The first term in Eq.(\ref{MRF}) is the data term which enforces the similarity between the corresponding positions of the high resolution depth map and the low resolution depth map. The second term in Eq.(\ref{MRF}) is the smoothness term which enforces the smoothness in the neighboring area. The data term and the smoothness term are balanced by the parameter $\alpha$. $\omega_{i,j}^{c}$ is defined as follows:
\begin{small}
\begin{equation}\label{ColorWeight}
    \omega _{i,j}^{c}=\exp\left(-\frac{\sum\limits_{k\in C}{|I_{i}^{k}-I_{j}^{k}{{|}^{2}}}}{3\times2\sigma _{c}^{2}}\right).
\end{equation}
\end{small}
where $C=\left\{ R,G,B \right\}$ represents the different channels of the color image. ${{\sigma }_{c}}$ is a constant defined by the user.

This framework has the following two constrains:

First, it is sensitive to the noise in the depth map. To clearly show this constrain, we take the derivative of the objective function in Eq.(\ref{MRF}) with respect to $D$ and let it equal to zero, then we can form an iterative formulation as:
\begin{small}
\begin{equation}\label{MRFUpdate}
    D_{i}^{n+1}=\frac{(1-\alpha)D_{i}^{0}+2\alpha \sum\limits_{j\in N(i)}{\omega _{i,j}^{c}D_{j}^{n}}}{(1-\alpha)+2\alpha \sum\limits_{j\in N(i)}{\omega _{i,j}^{c}}},\ i\in \Omega
\end{equation}
\end{small}

It is seen that Eq.(\ref{MRFUpdate}) has two important terms: the first term of the numerator is related to the data term in Eq.(\ref{MRF}) and the second term in Eq.(\ref{MRFUpdate}) is related to the smoothness term in Eq.(\ref{MRF}). The denominator in Eq.(\ref{MRFUpdate}) can be considered as a constant for a given $i\in \Omega $. It is irrelevant to the depth change. It is therefore skipped in our analysis. The first term contains only the initial value at the corresponding position. For a noise (including sensing noise and the noise caused by bicubic interpolation operation) free input, the initial value is close to the groundtruth. However, when the input contains significant noise, the initial value may be far away from the groundtruth. Simply adding this noisy initial value will greatly perturb the upsampling result. This implies that the data term in Eq.(\ref{MRF}) is sensitive to the noise. From the data term in Eq.(\ref{MRF}), it is seen that, during the upsampling process, the validity and the quality of depth value at each depth position are measured pixel-by-pixel in order to enforce the similarity between the the corresponding positions of the high resolution depth map and the low resolution depth map. Also, L2 norm is applied to calculate the similarity. From the analysis above, these two factors together make the data term sensitive to the noise.

Second, from the smoothness term in Eq.(\ref{MRF}), it is seen that the assumption of the co-occurrence of the color edge and the depth discontinuity is unnecessarily enforced. Consequently, the weighting value in this term (see Eq.(\ref{MRFUpdate})) is determined by the color difference of the corresponding positions on the color image. However, this assumption does not always well hold. Sometimes the locations with small depth difference on the depth map may contain large color difference on the color image. In this case, the obtained upsampled depth map may suffer from the texture copy artifacts. Another case is that the positions with large depth difference may contain small color difference on the color image. In such case, the second term will be close to the mean of the depth value in $N(i)$ which will result in blurring depth discontinuities.

The work in \cite{park2011high} extended the MRF by adding a more regularization term called the non-local means term. Moreover, unlike the weight in Eq.(\ref{ColorWeight}) which is only based on the color image, they further combine segmentation, color information, and edge saliency as well as the bicubic upsampled depth map to define the weight. Although the work in \cite{park2011high} somehow improves the performance when compared with \cite{diebel2005application}, it does not upgrade these two terms significantly and thus the performance improvement is limited. Figure \ref{SyntheticComparison} illustrates the $4\times$ upsampling results of a synthetic depth map\footnote{The result of the non-local means MRF in \cite{park2011high} is obtained with the default parameters in their own implementation.}. In this synthetic experiment, we show that these two methods are not robust against the noise and cannot handle the case where the color edge and the depth discontinuity are not consistent. We will further illustrate the texture copy artifacts of these two methods in the experimental part.

\section{The method}
\label{SecOurMethod}

In this section, we describe our optimization framework for upsampling the noisy low resolution depth map to a higher resolution one given a companion color image. Different from the previous work, we do not necessarily assume the co-occurrence of the color edge and the depth discontinuity. Meanwhile, it is more robust against the noise.

\subsection{The upsampling model}
\label{SecUpsamplingModel}

Our upsampling model consists of two terms: the data term and the smoothness term. Given a noisy low resolution depth map $D^L$, it is first interpolated to $D^0$ by bicubic interpolation. Then our upsampling model is formulated as:
\begin{small}
\begin{equation}\label{MyModel}
    D^H=\underset{D}{\mathop{\arg \min}}\,\left\{( 1-\alpha){{E}_{D}}(D,{{D}^{0}})+\alpha {{E}_{S}}(D)\right\}
\end{equation}
\end{small}
where ${{E}_{D}}\left( D,{{D}^{0}} \right)$ is the data term that makes the result to be consistent with the input. ${{E}_{S}}\left( D \right)$ is the smoothness term that reflects prior knowledge of the smoothness of our solution. These two terms are balanced with the parameter $\alpha $.\\
\textbf{The data term ${{E}_{D}}\left( D,{{D}^{0}} \right)$}: the data term is defined as:
\begin{small}
\begin{equation}\label{MyDataTerm}
    {{E}_{D}}\left( D,{{D}^{0}} \right)=\sum\limits_{i\in \Omega }{\sum\limits_{j\in N\left( i \right)}{{{\omega }_{i,j}}{{\varphi }_{D}}(|{{D}_{i}}-D_{j}^{0}|^2)}}
\end{equation}
\end{small}
where $\omega_{i,j}$ is a normalized Gaussian window that decreases the influence of ${\varphi}_{D}(|{D}_{i}-D_{j}^{0}|^2)$ where $j$ is far from $i$:
\begin{small}
\begin{equation}\label{SpatialWeight}
    {{\omega }_{i,j}}=\frac{1}{{{Z}_{i}}}\exp \left( -\frac{|i-j{{|}^{2}}}{2\sigma _{s}^{2}} \right),
    s.t.\ {{Z}_{i}}=\sum\limits_{j\in N\left( i \right)}{\exp \left( -\frac{|i-j{{|}^{2}}}{2\sigma _{s}^{2}} \right)}
\end{equation}
\end{small}
where ${{\sigma }_{s}}$ is a constant that is defined by the user. ${\varphi}_{D}(\cdot)$ is the robust error norm function that we denote as the exponential error norm. This function has long been used in image denoising such as local mode filtering \cite{van2001local} and nonlocal filtering \cite{pizarro2010generalised}. It is defined as:
\begin{small}
\begin{equation}\label{ErroNormFunction}
    {{\varphi}_{D}}(x^2)=2{{\lambda }^{2}}\left( 1-\exp \left( -\frac{{{x}^{2}}}{2{{\lambda }^{2}}} \right) \right)
\end{equation}
\end{small}

\begin{figure}
\centering
  \includegraphics[width=0.7\linewidth]{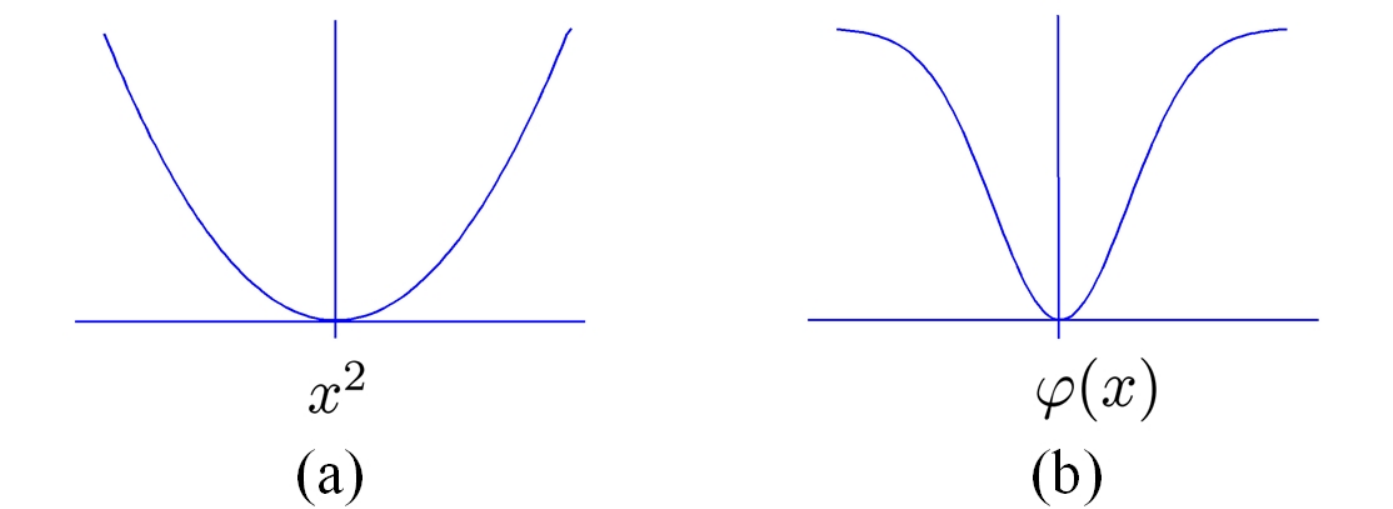}\\
  \caption{Illustration of two error norm functions. (a) The L2 error norm. (b) The exponential error norm proposed in this paper.}\label{ErrorNormFunctionComparision}
\end{figure}

The proposed exponential error norm is illustrated in Figure \ref{ErrorNormFunctionComparision}(b) to have a comparison with the L2 norm illustrated in Figure \ref{ErrorNormFunctionComparision}(a).

Our data term is different from that of previous methods \cite{diebel2005application}\cite{ferstl2013image}\cite{park2011high}\cite{yang2012depth}\cite{yang2014depth_recovery} which only use pixel-by-pixel depth difference modeled with L2 error norm in the data term. In Eq.(\ref{MyModel}), it can be seen that the new data term measures the depth differences pixel-by-patch by taking each reference pixel's neighboring region into account. According to \cite{ferstl2013image}, depth map is quite piece-wise smooth and thus pixel-by-patch difference calculation is more robust to the noise than pixel-by-pixel depth difference calculation. Such calculation is further normalized by Gaussian window $\omega_{i,j}$ in order to better maintain the local depth similarity in the neighboring area. Then, the pixel-by-patch depth difference is further modeled with the exponential error norm defined in Eq.(\ref{ErroNormFunction}) which is quite robust against the outliers \cite{pizarro2010generalised}. These together make the data term robust against the noise and will be further theoretically explained in Section \ref{SecAdvantageOverMRF}.\\
\textbf{The Smoothness term ${{E}_{S}}(D)$}: the smoothness term is guided by the companion color image. It is defined as:
\begin{small}
\begin{equation}\label{SmoothnessTerm}
    {{E}_{S}}\left( D \right)=\sum\limits_{i\in \Omega }{\sum\limits_{j\in N(i)}{{{{\tilde{\omega }}}_{i,j}}{{\varphi }_{S}}\left( |{{D}_{i}}-{{D}_{j}}|^2 \right)}}
\end{equation}
\end{small}
We adopt the same function in Eq.(\ref{ErroNormFunction}) to model the smoothness term, i.e. $\varphi _S(\cdot)=\varphi_D(\cdot)$. The color guided weight ${{\tilde{\omega }}_{i,j}}$ is defined as:
\begin{small}
\begin{equation}\label{ColorSpatialWeight}
    {{\tilde{\omega }}_{i,j}}={{\omega }_{i,j}^c}\cdot {\omega}_{i,j}
\end{equation}
\end{small}
where ${{\omega }_{i,j}}$ is the same as Eq.(\ref{SpatialWeight}) and ${\omega}_{i,j}^{c}$ is the same as Eq.(\ref{ColorWeight}).

Note that the color guided weight in our model is quite similar as that of the MRF in \cite{diebel2005application} except that there is an additional Gaussian window $\omega_{i,j}$. However, the MRF in \cite{diebel2005application} enforces the co-occurrence of the color edge and the depth discontinuity through Eq.(\ref{ColorWeight}). Depth discontinuity cues are completely based on color image in their work (as seen in Eq.(\ref{MRFUpdate})). Instead, the smoothness term (i.e. Eq.(\ref{SmoothnessTerm})) in the proposed framework relaxes such strict dependence due to $\varphi_S(\cdot)$. As we will further show in Section \ref{SecAdvantageOverMRF}, the smoothness term will result in a color image guided bilateral filter on the depth map at each update. Depth discontinuity cues are not only based on the color image but also based on the depth map itself during the optimization process. It is more capable of the case where the color edge and the depth discontinuity are not consistent with each other. This property will be the key element that makes our model tackle the texture copy artifacts and blur of depth discontinuities. We will further theoretically explain these in Section \ref{SecAdvantageOverMRF}.
%

\subsection{Further analysis of our model}
\label{SecAdvantageOverMRF}

We further analyze our upsampling model in order to demonstrate its advantages. By taking the derivative of the objective function of our model in Eq.(\ref{MyModel}) with respect to $D$, let it equal to zero and form the iterative formulation as:
\begin{small}
\begin{equation}\label{MyUpdate}
\begin{split}
    &D_{i}^{n+1}=\\
    &\frac{\left( 1-\alpha  \right)\sum\limits_{j\in N\left( i \right)}{{{\omega }_{i,j}}{{d}^n_{i,j}}}D_{j}^{0}+2\alpha \sum\limits_{j\in N\left( i \right)}{{{{\tilde{\omega }}}_{i,j}}{{s}^n_{i,j}}D_{j}^{n}}}{\left( 1-\alpha  \right)\sum\limits_{j\in N\left( i \right)}{{{\omega }_{i,j}}{{d}^n_{i,j}}}+2\alpha \sum\limits_{j\in N\left( i \right)}{{{{\tilde{\omega }}}_{i,j}}{{s}^n_{i,j}}}},\ i\in \Omega
\end{split}
\end{equation}
\end{small}
where we define
\begin{small}
\begin{equation}\label{ErrorNormFunctionDerivative}
\begin{split}
    &d^n_{i,j}={\varphi}'_{D}(|D^n_i - D^0_j|^2),\ \
    s^n_{i,j}={\varphi}'_{S}(|D^n_i - D^n_j|^2),\\
    &\ \ \ \ \ \ \ \ \ \ \ {\varphi}'_{D}(x^2)={\varphi}'_{S}(x^2)=\exp\left(-\frac{x^2}{2\lambda^2}\right)
\end{split}
\end{equation}
\end{small}
${{{\varphi }'}_{D}}(x^2)={{{\varphi }'}_{S}}\left(x^2 \right)$ is the derivative of ${{\varphi }_{D}}\left( x^2 \right)={{\varphi }_{S}}\left( x^2 \right)$ defined in Eq.(\ref{ErroNormFunction}).

Our method has the following advantages:

First, it is more robust against the noise in the input. The first term is the weighted sum of the initial depth values in $N(i)$. The weights are based on the difference of the latest updated depth map and the initial depth map, which is further normalized by the Gaussian window $\omega_{i,j}$. This term is related to the data term in the proposed framework. Compared with Eq.(\ref{MRFUpdate}), it can be shown that, in the newly proposed data term, the accuracy of upsampling in each round at each pixel is not affected by the initial upsampling value of the reference pixel only. Instead, the accuracy of upsampling in each round is based on the local measurement in its neighboring area $N(i)$. This results from the proposed pixel-by-patch depth difference measurement and is more robust against the noise. This effort is further enhanced by introducing $d^n_{i,j}$ which results from the proposed exponential error norm. When the original depth map contains significant noise, the depth value in the neighboring area is unlikely smooth which results in large $|D^n_i-D^0_j|^2$. This further causes the decrease of  $d^n_i$. Finally, such noisy area will have less effects on $D^n_i$ in each round of upsampling. Consequently, such data term is more robust to the noise. To our best understanding, it is a brand new data term used in the optimization framework for image guided depth upsampling.

The second term in the numerator in Eq.(\ref{MyUpdate}) is related to the smoothness term. Compared with the counterpart in Eq.(\ref{MRFUpdate}), it is shown that the existing methods enforce the assumption on the consistence between the color edge and the depth discontinuity, where the weight $\omega_{i,j}$ is only determined by the color difference in the neighboring area on the color image. In Eq.(\ref{MyUpdate}), it is shown that we relax such strict assumption by extending color image guided bilateral filtering onto each round of upsampling depth map $D^n$. The weight value is determined by three factors: 1) the color similarity in the local area (same as the existing method); 2) the distance between the reference pixel and its neighboring pixel (i.e. $\omega_{i,j}$); 3) the difference of depth value between the reference pixel and its neighboring pixels (i.e. $s^n_{i,j}$). $s_{i,j}^n$ reflects the property of the depth map. For the case where the depth region is homogeneous but the color information is not smooth at the corresponding area, $s^n_{i,j}$ can eliminate the efforts caused by $\omega^c_{i,j}$. Thus, it reduces the texture copy artifacts. For the case where depth region contains depth discontinuity but the color is smooth at the corresponding area on the color image, the second term will be close to a bilateral filter where $s^n_{i,j}$ well preserves the depth discontinuity. Thus the proposed smoothness term is more capable of cases where the color edge is not consistent with the depth discontinuity. We do not assume the color edge and the depth discontinuity to be necessarily consistent with each other.

Figure \ref{SyntheticComparison}(g) shows the result by our method. It is clear that the noise in the homogeneous regions have been well smoothed and the depth discontinuities can be properly preserved even there is no color edge corresponding to the depth discontinuity on the depth map.

\subsection{Data driven adaptive bandwidth selection}
\label{SecAdaptiveBandwidthSelectionModel}

In this paper, the parameter $\lambda$ is denoted as the \emph{bandwidth} of the exponential error norm function in Eq.(\ref{ErroNormFunction}) which is similar to the function of the tone term in the bilateral filter \cite{tomasi1998bilateral}. A large $\lambda $ has better noise smoothing but may over smooth the depth discontinuities. A small $\lambda $ can better preserve the depth discontinuities but performs poorly in noise smoothing. In this section, we describe an optimization method that adapts $\lambda $ to each pixel on the depth map. It is a data driven adaptive bandwidth selection. Because the depth map is quite piece wise smooth, we assume that the bandwidth is also regular and smooth. We add another regularization term that consists of the L2 norm of the gradient of ${\lambda}_{i}(i\in\Omega)$ to the objective function in Eq.(\ref{MyModel}). That is:
\begin{small}
\begin{equation}\label{BandwidthModel}
    E=\left( 1-\alpha  \right){{E}_{D}}\left( D,{{D}^{0}} \right)+\alpha {{E}_{S}}\left( D \right)+\beta \sum\limits_{i\in \Omega }{|\nabla {{\lambda }_{i}}{{|}^{2}}},\ i\in \Omega
\end{equation}
\end{small}
By minimizing Eq.(\ref{BandwidthModel}) with respect to ${{\lambda }_{i}}$ through the steepest gradient descent according to:
\begin{small}
\begin{equation}\label{BandwidthUpdate}
    \lambda _{i}^{n+1}=\lambda _{i}^{n}-\tau \frac{\partial E}{\partial \lambda _{i}^{n}},\ i\in \Omega
\end{equation}
\end{small}
where $\tau $ is the given updating rate and the derivative of the objective function is given by
\begin{small}
\begin{equation}\label{BandwidthDerivative}
\begin{split}
    &\frac{\partial E}{\partial \lambda^n_i} = \\
    &( 1-\alpha)\sum\limits_{j\in N(i)}{{{\omega }_{i,j}}\left[ 4{{\lambda }_{i}^n}\left( 1-{{d}_{i,j}^n} \right)-\frac{2{{\left( {{D}^n_{i}}-D_{j}^{0} \right)}^{2}}}{{{\lambda}_{i}^n}}{{d}^n_{i,j}} \right]}+\\
    &\alpha \sum\limits_{j\in N\left( i \right)}{{{{\tilde{\omega }}}_{i,j}}\left[ 4{{\lambda}_{i}^n}\left( 1-{{s}^n_{i,j}} \right)-\frac{2{{\left( {{D}^n_{i}}-{{D}^n_{j}} \right)}^{2}}}{{{\lambda }_{i}^n}}{{s}^n_{i,j}} \right]}+2\beta \sum\limits_{i\in \Omega }{\Delta {{\lambda }_{i}^n}}
\end{split}
\end{equation}
\end{small}
where ${{d}_{i,j}^n}$ and ${{s}_{i,j}^n}$ are the same as Eq.(\ref{ErrorNormFunctionDerivative}).

Depth map upsampling and the bandwidth selection are addressed in an iterative way through alternating the bandwidth update in Eq.(\ref{BandwidthUpdate}) and the depth map update in Eq.(\ref{MyUpdate}). Note that most components in Eq.(\ref{BandwidthDerivative}) have been already computed in Eq.(\ref{MyUpdate}). The computation cost for $\Delta\lambda$ is also quite small. So the computation cost of the bandwidth selection step is quite small indeed.

Figure \ref{CoverFigure}(f) illustrates a bandwidth map obtained by our method where dark regions correspond to smaller bandwidth values and bright regions correspond to larger bandwidth values. It is clear that this bandwidth map well corresponds to the character of the depth map shown in Figure \ref{CoverFigure}(b).
\section{Experiments}
\label{SecExperiments}

In this section, we perform the quantitative and qualitative evaluation of our method on both the synthetic data and real data. Comparison are performed with other state-of-art methods where the source codes are available. We show that our method can outperform other methods in most cases both quantitatively and qualitatively. For more experimental results, please see our supplementary material.

\subsection{Experiments on the synthetic data}
\label{SecSimulatedExperiments}

We first test our method on the synthetic data form the Middleburry dataset \cite{scharstein2007learning}. We reuse the data in \cite{yang2014depth_recovery} and compare our method with the MRF in \cite{diebel2005application} and its extension non-local means MRF in \cite{park2011high}, the image guided anisotropic total generalized variation upsampling in \cite{ferstl2013image}, the joint geodesic upsampling in \cite{liu2013joint}, the color guided auto-regression upsampling in \cite{yang2014depth_recovery} and the cross-based local multipoint filter in \cite{lu2012cross}. The upsamling results are evaluated in root mean square error (RMSE) between the original depth map and the upsampling result. Four upsampling factors are tested: $2\times /4\times /8\times /16\times $. All the values of both the color image and the depth map are normalized into interval $[0, 1]$ for convenience. However, the RMSE is still reported in terms that all the values of the depth map are in interval $[0, 255]$. The parameters are set as follows: we slightly tune $\alpha $ for different upsampling factors which is also the strategy adopted by other methods such as \cite{ferstl2013image}\cite{yang2014depth_recovery}. It is chosen as $0.8/0.9/0.96/0.99$ for  $2\times /4\times /8\times /16\times $ upsampling. Perturbations on the other parameters marginally affect the final results. $\beta$ in Eq.(\ref{BandwidthModel}) is set to $0.3$. The neighborhood $N(i)$ is chosen as a $19\times 19$ square patch. $\sigma _s$ and $\sigma_c $in Eq.(\ref{ColorSpatialWeight}) is set to $9$ and $\frac{10}{255}$ respectively. The initial value of $\lambda $ in Eq.(\ref{BandwidthUpdate}) is set to $\frac{7}{255}$ for all $i\in\Omega$. Its updating rate $\tau$ in Eq.(\ref{BandwidthUpdate}) is $0.3$.

\begin{table*}
\centering
\caption{Quantitative comparison on the synthetic data from the Middlenurry dataset \cite{scharstein2007learning}. Four upsampling factors are tested, i.e. $2\times/4\times/8\times/16\times$. The results are evaluated in RMSE and the best results are in bold.}\label{SimulatedRMSE}

\resizebox{1\textwidth}{!}
{
\begin{tabular}{|c|cccc|cccc|cccc|cccc|cccc|cccc|}

\hline
  \multicolumn{1}{|c}{\multirow{2}{*}{}} & \multicolumn{4}{|c|}{\emph{Art}} & \multicolumn{4}{c|}{\emph{Book}} & \multicolumn{4}{c|}{\emph{Dolls}} & \multicolumn{4}{c|}{\emph{Laundry} } & \multicolumn{4}{c|}{\emph{Moebius}} & \multicolumn{4}{c|}{\emph{Reindeer}}\\

  \cline{2-25} 
  & $2\times$ & $4\times$ & $8\times$ & $16\times$ & $2\times$ & $4\times$ & $8\times$ & $16\times$ & $2\times$ & $4\times$ & $8\times$ & $16\times$ & $2\times$ & $4\times$ & $8\times$ & $16\times$ & $2\times$ & $4\times$ & $8\times$ & $16\times$ & $2\times$ & $4\times$ & $8\times$ & $16\times$ \\
  \hline

  CLMF \cite{lu2012cross} & 1.19 & 1.77 & 2.95 & 4.91 & 0.9 & 1.48 & 2.38 & 3.36 & 0.87 & 1.44 & 2.32 & 3.3 & 0.96 & 1.56 & 2.54 & 3.85 &	 0.94 & 1.55 & 2.5 & 3.81 & 0.96 &	1.54 & 2.37 & 3.25 \\

  JGF \cite{liu2013joint} & 2.36 & 2.74 &	3.64 & 5.46 & 2.12 & 2.25 &	2.49 & 3.25 & 2.09 & 2.24 &	2.56 & 3.28 & 2.18 & 2.4 & 2.89 & 3.94 & 2.16 & 2.37 & 2.85 &	3.9 & 2.09 & 2.22 &	2.49 & 3.25 \\

  MRF \cite{diebel2005application} & 1.24 & 1.69 & 2.51 & 3.99 & 0.74 & 1.04 & 1.53 & 2.3 & 0.75 & 1.04 & 1.5 & 2.19 & 0.78 & 1.12 & 1.67 & 2.73 & 0.79 & 1.08	& 1.57 & 2.33 & 0.83 & 1.11 & 1.65 & 2.57 \\

  NLM-MRF \cite{park2011high} & 1.66 & 2.47 & 3.44 & 5.55 & 1.19 & 1.47 & 2.06 & 3.1 & 1.19 & 1.56 & 2.15 & 3.04 & 1.34 & 1.73 & 2.41 & 3.85 & 1.2 & 1.5 & 2.13 & 2.95 & 1.26 & 1.65 & 2.46 & 3.66 \\

  TGV \cite{ferstl2013image} & 0.8 & 1.21 & 2.01 & 4.59 & 0.61 & 0.88 & 1.21 & 2.19 & 0.66 & 0.96 & 1.38 & 2.88 & 0.61 & 0.87 & 1.36 & 3.06 & 0.57 & \textbf{0.77} & 1.23 & 2.74 & 0.61 & 0.85 & 1.3 & 3.41 \\

  AR \cite{yang2014depth_recovery} & 0.92 & 1.23 & 2.1 & 3.9 & 0.74 & 0.85 & 1.23 & 1.96 & 0.8 & 0.97 & 1.35 & 2.24 & 0.73 & 0.93 & 1.34 & 2.24 & 0.72 & 0.82 & 1.25 & 2.16 & 0.77 & 0.87 & 1.3 & 2.73 \\

  Ours & \textbf{0.69} & \textbf{1.14} & \textbf{1.89} & \textbf{3.26} & \textbf{0.56} & \textbf{0.8} & \textbf{1.16} & \textbf{1.72} & \textbf{0.63} & \textbf{0.88} & \textbf{1.23} & \textbf{1.78} & \textbf{0.6} & \textbf{0.84} & \textbf{1.22} & \textbf{2.13} & \textbf{0.55} & 0.78	& \textbf{1.14} & \textbf{1.74} & \textbf{0.58} & \textbf{0.84} & \textbf{1.27} & \textbf{2.13} \\

  \hline
\end{tabular}
}
\end{table*}

Table \ref{SimulatedRMSE} summarizes the quantitative comparison results on the Middleburry dataset \cite{scharstein2007learning}. It is clear that our method outperforms other methods in almost all the cases. Especially, results by our method have a great improvement over the ones by the MRF in \cite{diebel2005application} and the non-local means MRF in \cite{park2011high}. Even when compared with the most recent proposed methods in \cite{ferstl2013image} and \cite{yang2014depth_recovery}, our method have smaller RMSE in almost all the cases. Note that the method in \cite{ferstl2013image} needs thousands of iterations to converge which is quite time consuming, namely $10,000$ iterations for $2\times$ and $4\times$ upsampling, $15,000$ iterations for $8\times$ upsampling and $30,000$ iterations for $16\times$ upsampling which needs hours with their own implementation\footnote{we perform the comparison with their own MATLAB implementation and all the parameters recommended by the author. The computational time reported in their paper was implemented with GPU acceleration}. However, our method only needs a few iterations to converge, for example, less than 5 iterations for $2\times$ upsampling and less than 50 iterations for $16\times$ upsampling which only needs very few minutes. The computation cost of the method in \cite{yang2014depth_recovery} is also quite huge due to its complex shape-based color guided weights. Our method is also a magnitude faster than theirs. Also, our computational cost doubles the MRF in \cite{diebel2005application} but is about half of the non-local MRF in \cite{park2011high}.

\begin{figure}
\centering
  \includegraphics[width=1\linewidth]{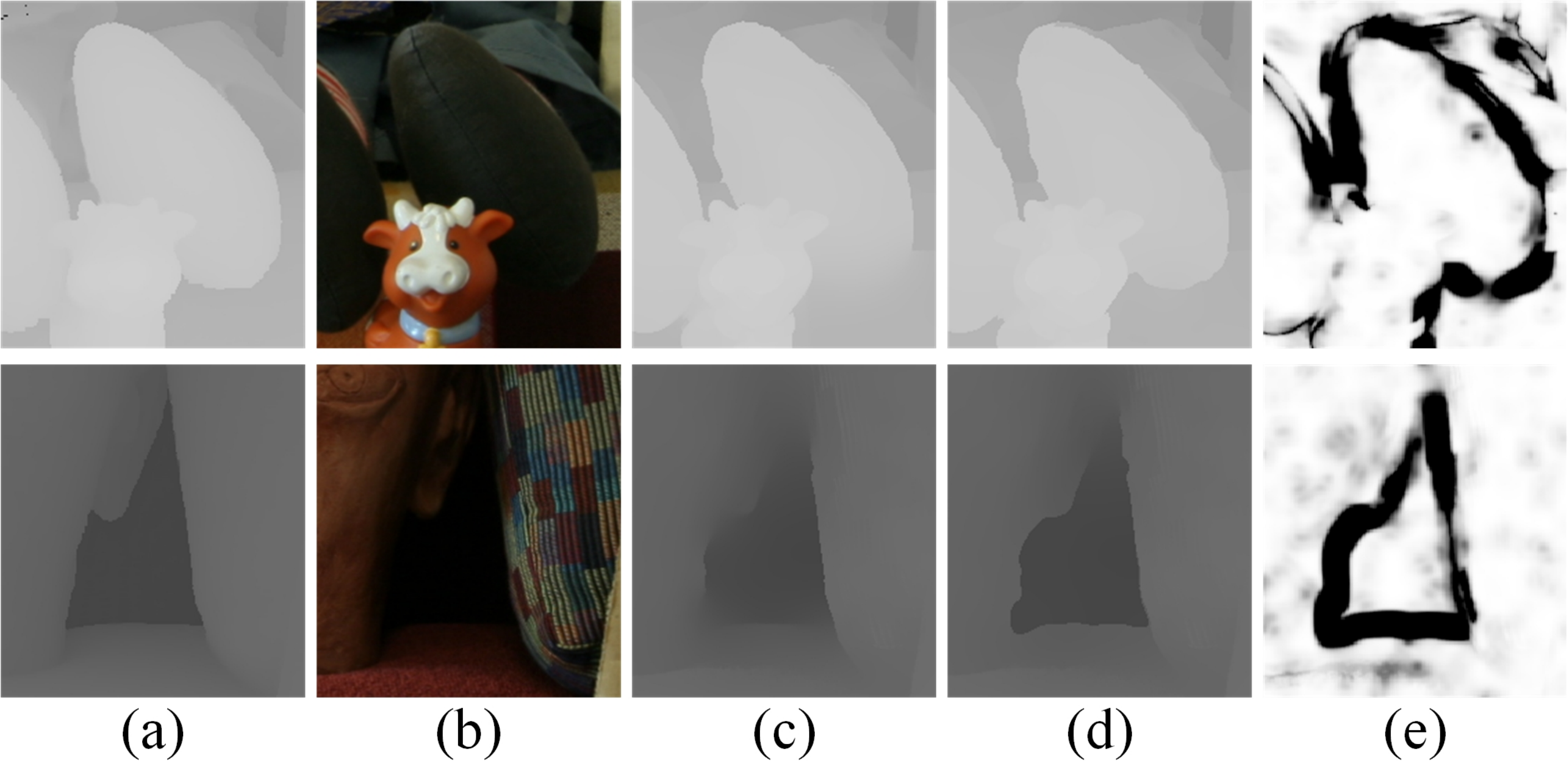}\\
  \caption{Visual comparison of our method for $8\times$ upsampling with and without bandwidth selection. The first row are patches from \emph{Dolls} and the second row are patches from \emph{Reindeer}. (a) The groundtruth. (b) The corresponding color image. (c) Results obtained without bandwidth selection. (d) Results obtained with bandwidth selection and (e) the corresponding bandwidth maps.}\label{BandwidthSelectionComparison}
\end{figure}

Figure \ref{BandwidthSelectionComparison} illustrates the improvement of upsampling results by our bandwidth selection. Without bandwidth selection, it is quite easy to smooth the depth discontinuities where the corresponding color edges are weak. Moreover, our bandwidth selection can further help to preserve the fine details in the depth map. Figure \ref{CoverFigure}(e) clearly illustrates the improvement. The small holes in the ring are properly preserved with the adaptive bandwidth selection while this cannot be achieved without the adaptive bandwidth selection as shown in Figure \ref{CoverFigure}(d).

\begin{figure*}[!h]
\centering
  \includegraphics[width=1\linewidth]{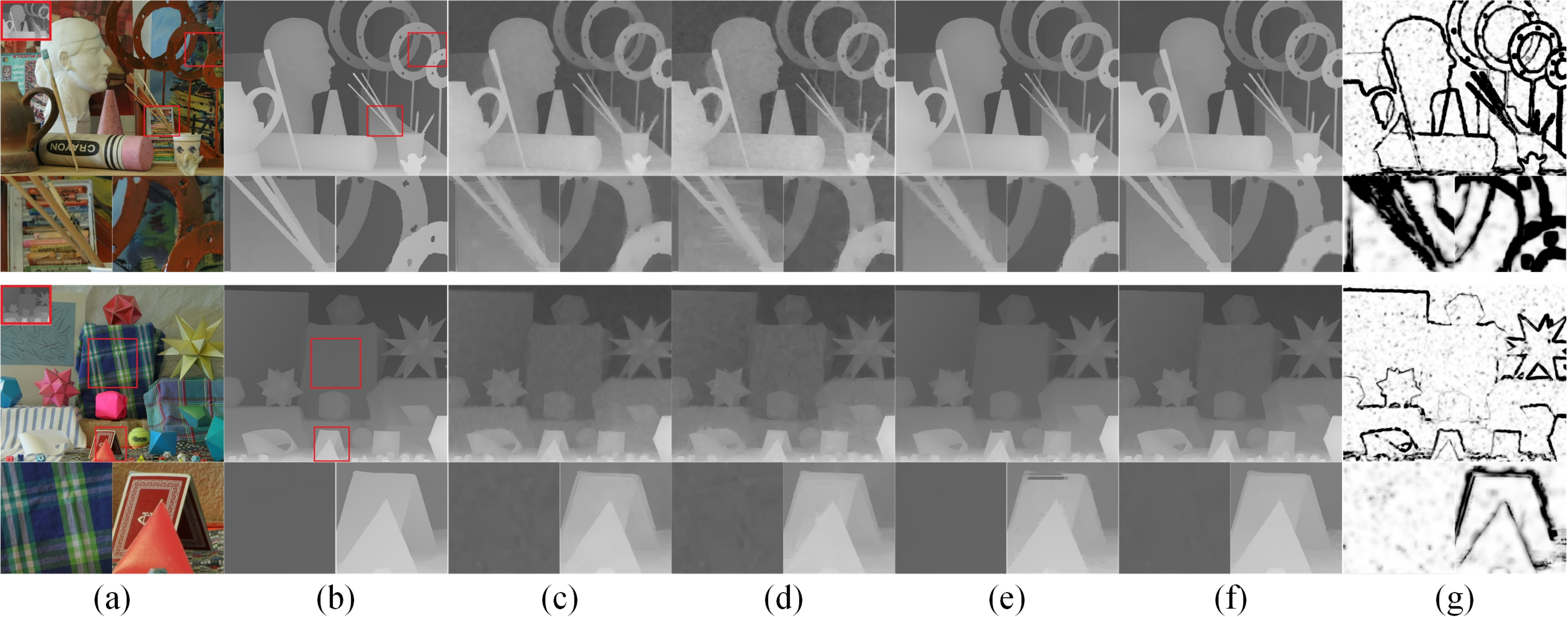}\\
  \caption{Visual comparison of $8\times$ upsampling results of \emph{Art} and \emph{Moebius} from the Middleburry dataset \cite{scharstein2007learning}. (a) The input depth maps (in red boxes) and the corresponding color images. (b) The  groundtruth depth maps. The results obtained by (c) the MRF in \cite{diebel2005application}, (d) its extension non-local means MRF in \cite{park2011high}, (e) the image guided total generalized variation upsampling in \cite{ferstl2013image}, (f) our method and (g) the corresponding bandwidth maps by our bandwidth selection. The regions in red boxes which contain fine details are highlighted.}\label{8UpsamplingComparison}
\end{figure*}

\begin{figure*}[!h]
\centering
  \includegraphics[width=1\linewidth]{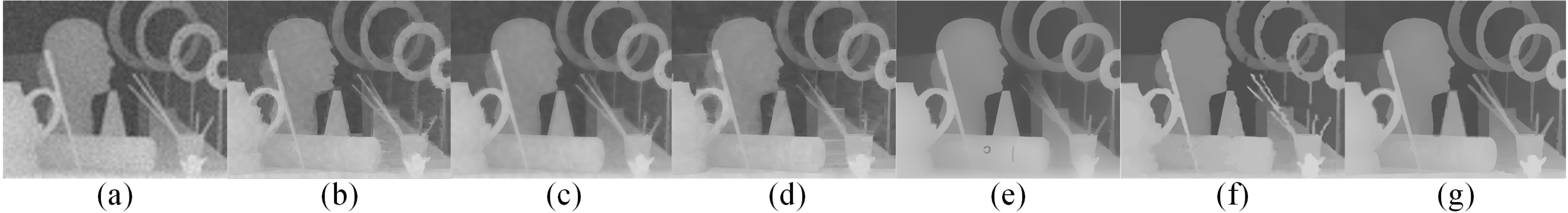}\\
  \caption{Visual comparison of $16\times$ upsampling results of \emph{Art} from the Middleburry dataset \cite{scharstein2007learning}. (a) The bicubic interpolated depth map, and results by (b) the joint geodesic upsampling in \cite{liu2013joint}, (c) the MRF in \cite{diebel2005application}, (d) the non-local means MRF in \cite{park2011high}, (e) the image guided generalized total variation upsampling in \cite{ferstl2013image} (f) the color guided auto-regression upsampling in \cite{yang2014depth_recovery} and (g) our method.} \label{16UpsamplingComparison}
\end{figure*}

Figure \ref{8UpsamplingComparison} shows the examples of $8\times$ upsampling for visual comparison. It is clear that results by the MRF in \cite{diebel2005application} and the non-local means MRF in \cite{park2011high} cannot well smooth the noise in homogeneous regions and properly preserve the depth discontinuities. As shown in the highlighted region of the \emph{Moebius}, results by their methods suffer from heavy texture copy effect while ours is not. The method in \cite{ferstl2013image} can avoid the texture copy effect, but it also cannot preserve the fine details, for example, the small holes on the ring of \emph{Art} and the small object of \emph{Moebius} which are shown in the highlighted regions. However, our method can not only well smooth the noise in the homogeneous regions and avoid texture copy but also properly preserve the depth discontinuities even the fine details. Figure \ref{16UpsamplingComparison} further shows $16\times$ upsampling results comparison. It is clear that even for such a large upsampling factor, our method can still well smooth the noise and preserve sharp depth discontinuities while none of the compared methods could yield such performance as clearly shown in the figure.

\begin{table*}[!h]
\centering
\caption{Quantitative comparison on real data from \cite{ferstl2013image}. The error is calculated as RMSE to the measured
groundtruth in mm. The best results are in bold.}\label{RealRMSE}

\resizebox{0.9\textwidth}{!}
{
\begin{tabular}{|c|ccccccccc|}
  \hline
   & Bicubic & CLMF \cite{lu2012cross} & JGF \cite{liu2013joint} & JBF \cite{kopf2007joint} & MRF \cite{diebel2005application} & NLM-MRF \cite{park2011high} & TGV \cite{ferstl2013image} & AR \cite{yang2014depth_recovery} & Ours \\
  \hline
  \emph{Books} & 16.23mm &  13.89mm & 17.39mm & 15.42mm & 13.87mm & 14.31mm & 12.8mm & 13.28mm & \textbf{12.58mm} \\
  \emph{Devil} & 17.78mm &  14.55mm & 19.02mm & 16.47mm & 15.36mm & 15.36mm & 14.97mm & 14.73mm & \textbf{14.28mm} \\
  \emph{Shark} & 16.66mm &  15.1mm & 18.17mm & 17.16mm & 16.07mm & 15.88mm & 15.53mm & 15.86mm & \textbf{14.74mm} \\
  \hline
\end{tabular}
}
\end{table*}


\begin{figure*}[!h]
\centering
  \includegraphics[width=1\linewidth]{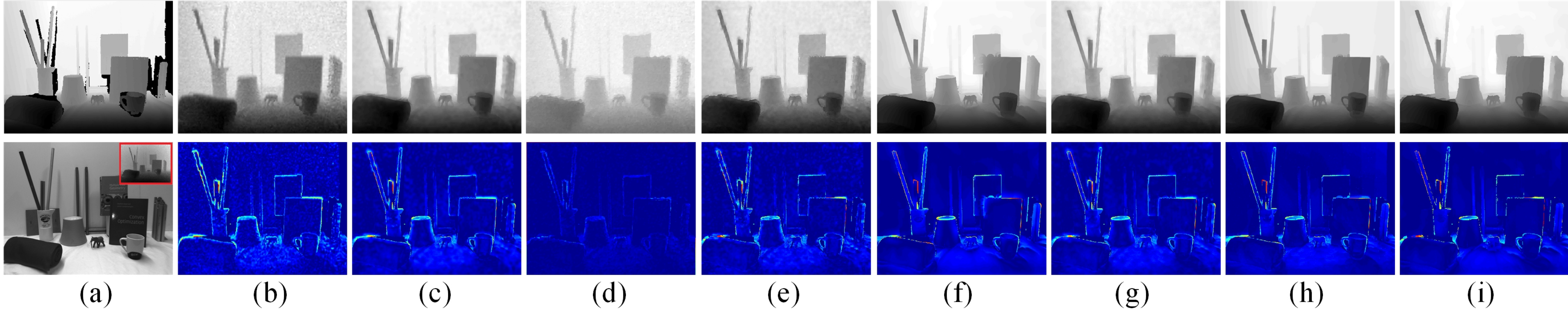}\\
  \caption{Visual comparison on the real data \emph{Books} from \cite{ferstl2013image}. The first row are (a) the measured groundtruth and results by (b) bicubic interpolation, (c) the cross-based local multipoint filter in \cite{lu2012cross}, (d) the joint geodesic upsampling in \cite{liu2013joint}, (e) the joint bilateral filter in \cite{kopf2007joint}, (f) the MRF in \cite{diebel2005application}, (g) the non-local means MRF in \cite{park2011high}, (h) the image guided total generalized variation upsampling in \cite{ferstl2013image} and (i) our method. The intensity image together with the input depth map (in the red box) and corresponding error maps are shown in the second row.}\label{RealDataComparison}
\end{figure*}

\subsection{Experiments on the real data}
\label{SecRealExperiments}

To test the effectiveness of our method on real ToF depth maps, we further test our method on the real ToF depth maps from \cite{ferstl2013image}. As far as we know, this is the only real ToF dataset that \emph{provides groundtruth}. Three depth maps are included in this dataset, namely \emph{Books}, \emph{Devil} and \emph{Shark}. All the values in the depth map are the real depth values from the camera to the measured objects (in mm). The upsampling factor is about $6.25\times$. Table \ref{RealRMSE} summarizes the quantitative comparison on this real dataset. Our method outperforms all the other methods on these three depth maps. Figure \ref{RealDataComparison} illustrates the visual comparison of the \emph{Books} upsampling results. Note that the result of the MRF \cite{diebel2005application} has strong texture copy effect at the up left corner of the book. The result of the non-local means MRF \cite{park2011high} still contains noise in the homogeneous regions. Our method can well smooth the noise in homogeneous regions and preserve sharp depth discontinuities.

\section{Conclusion}
\label{SecConclusion}

In this paper, we propose a novel method for depth map upsampling guided by the high resolution color image. We model the upsampling work with an optimization framework that consists of a brand new data term and a new smoothness term. The proposed data term is based on the pixel-patch difference and is modeled with an exponential error norm function. It has been proved to be more robust against the noise than the one based on pixel-pixel difference with L2 norm as the error norm function. We relax the too strict assumption on the consistency between the color edge and the depth discontinuity which is adopted by most existing methods. The new smoothness term makes our model not only obtain the depth discontinuity cues from the guided color image but also the depth map itself. This makes our model well tackle the texture copy artifacts and preserve sharp depth discontinuities even when there is no color edge correspondence. Moreover, a data driven scheme is proposed to adaptively select the proper bandwidth of the exponential error norm function. This helps to further improve the upsampling quality where fine details and sharp depth discontinuities could be preserved even for a large upsampling factor, $8\times$ and $16\times$ for example. Experimental results on both synthetic data and real data have shown our method outperforms other state-of-art methods.

{\small
{
\bibliographystyle{ieee}
\bibliography{egpaper_for_review}
}
}

\end{document}